\documentclass[11pt,a4paper]{article}
\usepackage{acl2015}
\usepackage{times}
\usepackage{url}
\usepackage{latexsym}
\usepackage{amssymb}
\usepackage{epsfig}

\usepackage{calc}
\newsavebox\CBox
\newcommand\hcancel[2][0.5pt]{%
  \ifmmode\sbox\CBox{$#2$}\else\sbox\CBox{#2}\fi%
  \makebox[0pt][l]{\usebox\CBox}%  
  \rule[0.5\ht\CBox-#1/2]{\wd\CBox}{#1}}

\newcommand{\ignore}[1]{}

\title{Replication and Generalization of PRECISE}

\author{Michael Minock and Nils Everling \\
TCS/CSC \\
  KTH Royal Institute of Technology, Stockholm, Sweden \\
  {\tt \{minock,everling\}@csc.kth.se}\\}

\date{}
\newtheorem{definition}{Definition}

\begin{document}
\maketitle
\begin{abstract} %200 words maximum.
  This report describes an initial replication study of the PRECISE
  system and develops a clearer, more formal description of the
  approach. Based on our evaluation, we conclude that the PRECISE
  results do not fully replicate. However the formalization developed
  here suggests a road map to further enhance and extend the approach
  pioneered by PRECISE.

  {\bf After a long, productive discussion with Ana-Maria Popescu (one
    of the authors of PRECISE) we got more clarity on the PRECISE
    approach and how the lexicon was authored for the GEO
    evaluation. Based on this we built a more direct implementation
    over a repaired formalism. Although our new evaluation is not yet
    complete, it is clear that the system is performing much better
    now. We will continue developing our ideas and implementation and
    generate a future report/publication that more accurately
    evaluates PRECISE like approaches.  }
\end{abstract}

\section{Introduction}

 It is no secret that the cost of configuring and maintaining natural
 language interfaces to databases is one of the main obstacles to
 their wider adoption\cite{androutsopoulos}.  While recent work has
 focused on learning approaches, there are less costly alternatives
 based on only lightly naming database elements (e.g.  relations,
 attributes, values) and reducing question interpretation to graph
 match \cite{meng99,precise03}.

 What is particularly compelling about PRECISE
 \cite{precise03,precise04} is the claim that for a large and well
 defined class of \emph{semantically tractable} questions, one can
 guarantee correct translation to SQL. Furthermore PRECISE leverages
 off-the-shelf open domain syntactic parsers to help guide query
 interpretation, thus requiring no tedious grammar
 configuration. Unfortunately after PRECISE was introduced there has
 not been much if any follow up.  This paper aims to evaluate these
 claims by implementing the model and conducting experiments
 equivalent those done by the designers of PRECISE.

\begin{figure}[!htb]
	\centering
		\includegraphics[width=0.4\textwidth]{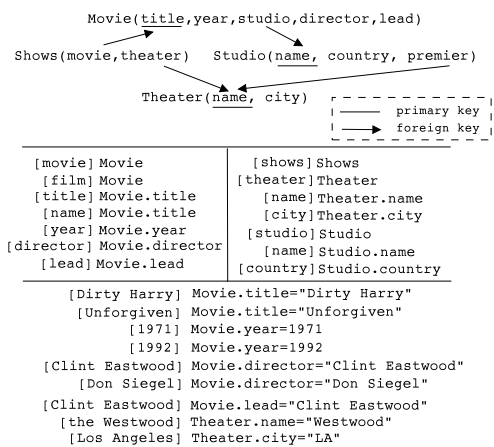}
	        \caption{Example schema and partial lexicon}
	\label{fig:db}
\end{figure}

 Consider the example database schema depicted at the top of figure
 \ref{fig:db}. Although this schema is small, it contains a
 many-to-many-relationship (movies to theaters) and a many-to-one
 relation from movie to studio. The schema is also cyclic (via foreign
 key-based joins) based on the somewhat contrived foreign key
 \verb+premier+ from \verb+Studio+ to \verb+Theater+ to indicate that
 a studio shows their premiers in a specific theater.

\section{A more `precise' formalization}

\subsection{The Database}

Databases are represented as a disjoint set of \emph{relations} $R$,
\emph{attributes} $A$ and \emph{values} $V$ which together are the
\emph{database elements} $E = R \cup A \cup V$. The function
$\mathsf{relOf}: A \rightarrow R$ and $\mathsf{attOf}:V \rightarrow A$
gives the relation of an attribute and the attribute of a value
respectively. The Boolean function $\mathsf{key}:A \rightarrow
\{true,false\}$ is true for attributes that are primary keys of their
corresponding relations.

\subsection{Words, phrases and the lexicon}

We consider $\mathcal{W}$ to be the set of \emph{words} in a natural
language and the set of \emph{phrases} $\mathcal{P}$ to be all finite
non-empty word sequences. We speak of $w_i$ being $i$-th word of the
phrase $p= [w_1,...,w_i,...,w_n]$,$p[i]=w_i$, and $|p|$ is the length
of $p$. $\mathsf{WH} = \{[who],$ $[which], [what], [where],$
$[when], [how]\} \subsetneq \mathcal{P}$ and $\mathsf{Stop} =
\{[are], [the],$ $[on], [a], [in], [is], [be], [of], [do], [with],
  [have],$ $[has]\}\subsetneq \mathcal{P}$.  Assume a special function
  $\mathsf{stem}: \mathcal{W} \rightarrow \mathcal{W}$ which stems
  words according to morphology of the natural language. The lexicon
  $\mathcal{L} \subsetneq \mathcal{P} \times E$ is a set of phases paired
  with database elements. See the bottom part of figure \ref{fig:db}
  for an example lexicon. Finally assume the function
  $\mathsf{compWH}: A \cup R \rightarrow 2^\mathsf{WH}$ which
  associates with every attribute and relation a set of compatible
  WH-words (e.g. $\mathsf{compWH}(Movie.name)= \{[which],[what]\}$).

\subsection{Assigning words to phrases}

A user question $q$ is a sequence of words $q=[w_1,...,w_n]$. An off
the shelf syntactic parser determines an \emph{attachment relation}
between words. Formally, $AW_q(i,j) \Leftrightarrow \{i,j\} \subseteq
\{1,..,n\} \wedge w_i \textsf{ attaches to } w_j$.

A \emph{covering assignment} $\zeta: \{1,..,n\} \rightarrow
\mathcal{P}_{\zeta} \cup \mathsf{Stop} \cup \mathsf{WH}$ observes the
following properties:

\begin{itemize}
\item[1.] (words belong to phrases)

  if $\zeta(i) = p_j$ then $(\exists e) ((p_j,e) \in
  \mathcal{L}) \vee p_j \in  \mathsf{Stop} \cup  \mathsf{WH}$

\item[2.] (phrases are complete)

  if $\zeta(i) = p_j$ and $i=1 \vee (\zeta(i-1) = p_k \wedge k
  \neq j$), then $(\forall m) ((m \in \mathbb{N})(m \ge 0) \wedge (m < |p_j|)
  \Rightarrow \mathsf{stem}(q[i+m]) = \mathsf{stem}(p_j[m]))$
\end{itemize}

The set of lexicon phrases in the range of $\zeta$ is
$\mathcal{P}_\zeta$. This corresponds to what the authors of PRECISE
call \emph{tokenization}.

\subsection{Mapping to database elements}

Consider $\phi_\zeta:\mathcal{P}_\zeta \rightarrow E$ to be an
injective function with image $E_{\phi_{\zeta}}$. This corresponds to
the \emph{matching process} in the PRECISE papers where each phrase is
paired uniquely with a database element.

We define a binary attachment relation $AE_{\phi_{\zeta}}$ on the
elements in $ E_{\phi_{\zeta}}$ which carries the attachment
information on words to attachment relations on elements. Formally,
$(\forall e_i)(\forall e_j)(AE_{\phi_{\zeta}}(e_i,e_j) \Leftrightarrow
\{e_i,e_j\} \subseteq E_{\phi_{\zeta}} \wedge (\exists w_{i'})(\exists
w_{j'})(\phi_\zeta (\zeta (i')) = e_i \wedge \phi_\zeta (\zeta (j')) =
e_j) \wedge AW_{q}(w_{i'},w_{j'}))$

A mapping that satisfies the following additional constraints is
\emph{valid}:

\begin{itemize}

\item[1.] (unique focus)
  
 $(\exists! e_{focus})(e_{focus} \in E_{\phi_{\zeta}} \cap (A \cup R))$ 
  
\item[2.] (necessary value correspondences)

  $(\forall e)(e \in E_{\phi_{\zeta}} \wedge e \in V \Rightarrow
  (\mathsf{attOf}(e) \in E_{\phi_{\zeta}} \wedge
  AE_{\phi_{\zeta}}(e,\mathsf{attOf}(e))) \vee
  (\mathsf{relOf}(\mathsf{attOf}(e)) \in E_{\phi_{\zeta}} \wedge
  AE_{\phi_{\zeta}}(e,\mathsf{relOf}(\mathsf{attOf}(e)))) \vee
  \underline{\mathsf{key}(\mathsf{attOf}(e))=true})$

\item[3.] (necessary attribute correspondences) 

$(\forall e)(e \in E_{\phi_{\zeta}} \wedge e \in A \wedge e \neq
  e_{focus} \Rightarrow
  ((\exists! e') (e' \in E_{\phi_{\zeta}} \wedge e' \in V \wedge
  \mathsf{attOf}(e') = e \wedge AE_{\phi_{\zeta}}(e,e')))$

\item[4.] (necessary relation correspondences)

  $\hcancel{(\forall
    e)(e \in E_{\phi_{\zeta}} \wedge e \in R \Rightarrow (\exists e')
    (e \in E_{\phi_{\zeta}}} \wedge$ $\hcancel{(e' \in A \wedge \mathsf{relOf}(e') = e)
    \vee (e' \in V \wedge}$ $\hcancel{\mathsf{attrOf}(\mathsf{relOf}(e')) = e)))}$
  
\end{itemize}

Property 1 states that there is a distinguished attribute or relation
that is the focus of the question. Property 2 states that values must
be paired with either an attribute (e.g. ``... title unforgiven
...''), or via ellipsis paired with a relation (e.g. ``... the movie
unforgiven''), or, if the value is a key itself, we have a highly
elliptical case where the value may stand on its own
(e.g. ``unforgiven''). Property 3 says that non-focus attributes must
pair with a value (e.g. in ``...movies of year 2000...'' 2000 serves
this role). Property 4 was included in the PRECISE papers, but we
found it unnecessary.

\subsection{Semantically tractable questions}

\begin{definition} (Semantically Tractable Question) For a given
  question $q$, lexicon $\mathcal{L}$ and attachment relation $AW_q$,
  $q$ is semantically tractable if there exists a covering assignment
  $\zeta$ over $q$ for which there is a valid mapping: $\phi_\zeta$
  and $\zeta$ assigns a word in $q$ to $\mathsf{WH}$ which is
  compatible with $e_{focus} \in E_{\phi_{\zeta}}$.
\end{definition}

\begin{definition} (Unambiguous Semantically Tractable Question)
 For a given question $q$, lexicon $\mathcal{L}$ and attachment
 relation $AW_q$, $q$ is unambiguous semantically tractable if $q$ is
 semantically tractable and $(\forall \zeta')(\forall \zeta'')(\forall
 \phi'_{\zeta'})(\forall \phi''_{\zeta''})(\phi'_{\zeta'} \textsf{ is
   valid } \wedge \phi''_{\zeta''} \textsf{ is valid } \Rightarrow
 E_{\phi'_{\zeta'}}=E_{\phi''_{\zeta''}})$
\end{definition}

\begin{figure}[!htb]
	\centering
		\includegraphics[width=0.4\textwidth]{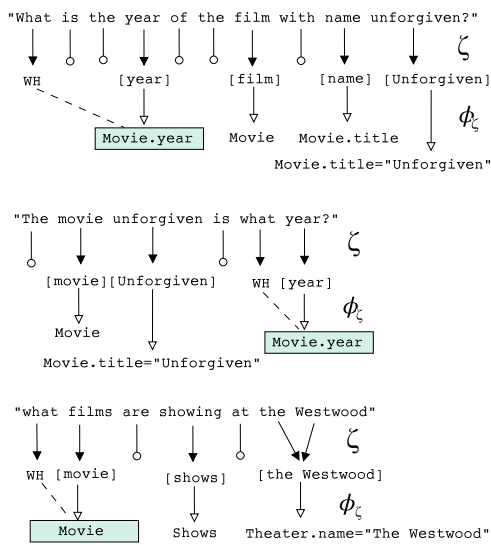}
	        \caption{Example valid mappings}
	\label{fig:valid}
\end{figure}

Figure \ref{fig:valid} shows three valid mappings given the schema and
lexicon in figure \ref{fig:db}. An additional example is ``what films
did Don Siegal direct with lead Clint Eastwood?'' This is a
\emph{unambiguous semantically tractable question} so long as `Don
Siegal' attaches to `direct' and not `lead', and `Clint Eastwood'
attaches to `lead' and not `direct'.

\subsection{Generating SQL}

The PRECISE papers say little about generating SQL from sets of
database elements. That said, it seems fairly straight forward. The
focus element becomes the attribute (or * in case focus is a relation)
in the SQL \verb+SELECT+ clause. All the involved or implied relation
elements are included in the \verb+FROM+ clause. The value elements
determine the simple equality conditions in the \verb+WHERE+
clause. Adding the join conditions is not formalized in PRECISE, but
we assume it means adding the minimal set of equality joins necessary
to span all relation elements. For cyclic schemas this can lead to
ambiguity. For example, while there is a unique valid mapping for the
question ``What movies at the Westwood'', join paths via \verb+studio+
or \verb+shows+ are possible in the schema of figure \ref{fig:db}.

\section{Our Implementation}

Our JAVA-based open-source
implementation\footnote{\url{https://github.com/everling/PRECISE}},
corresponds to the formal definition of section 2. Like PRECISE,
$\zeta$ assignments are computed via a brute force search and
candidate valid mappings $\phi_\zeta$ are solved for via reduction to
graph max-flow. Candidate solutions are filtered based on attachment
relations obtained from the Stanford Parser \cite{stanford}.  We
generate all possible SQL queries for all valid mappings.

\section{Our Evaluation}

Like the earlier work, we evaluated our system on {\sc
  Geoquery}\footnote{\url{www.cs.utexas.edu/users/ml/geo.html}}. Since
very little information has been disclosed regarding how PRECISE
purportedly handled \emph{superlatives} (``What is the most populous
city in America?''), \emph{aggregation} (``What is the average
population of cities in Ohio?''), and \emph{negation} (``Which states
do not border Kansas?''), we simply excluded these types of questions
from our evaluation. This reduced our tests to 442 (of 880) {\sc
  Geoquery} Questions.

In theory, PRECISE could be deployed immediately on any relational
database. However, we found the automatic approach to be very erratic,
generating many irrelevant synonyms. Part of speech-tagging (POS),
which can help to narrow down the senses of a word, is difficult to
determine automatically from database element names. Even with the
correct POS identified a word might have irrelevant senses which muddle
the lexicon. For example, WordNet has 26 noun senses of the word
“point” in the Geoquery attribute \verb+highlow.lowest_point+, one of
which has a synonym being `state'. Hence we decided to manually add
mappings to the lexicon.  Another reason to do this was to map
relevant phrases which would not have been generated automatically
otherwise. For example, to correctly answer the question ``What major
rivers are in Texas?'' the phrase \verb+[major river]+ had to be
associated with the relation \verb+river+. 

\begin{figure}[!htb]
	\centering
		\includegraphics[width=0.4\textwidth]{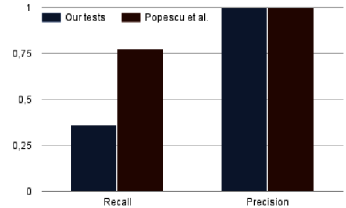}
                \vspace{-.3cm}
	\caption{A negative replication result}
	\label{fig:non-repeat}
\end{figure}

Out of these 448 questions, 162 were answered correctly by our
replication of PRECISE. This does not accord to previously published
recall results (see figure \ref{fig:non-repeat}). On the positive
side, there were no questions for which PRECISE returned a single
wrong query.

\begin{figure}[!htb]
	\centering
		\includegraphics[width=0.5\textwidth]{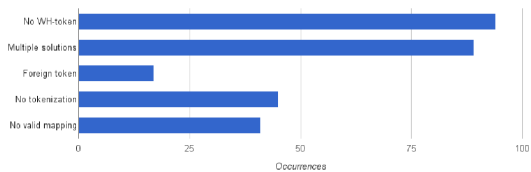}
	\caption{Sources of rejection}
	\label{fig:problems}
\end{figure}

Figure \ref{fig:problems} breaks down the reasons why the 286
remaining questions for were rejected by our system: 94 questions
contained no WH-word, 17 sentences contained non-stop words which the
lexicon did not recognize as part of any phrase, 45 questions had at
least one $\zeta$, but no $\phi_\zeta$ could be found that mapped
one-to-one and onto a set of elements, 41 questions had a $\phi_\zeta$
that was one-to-one and onto, but no valid mapping could be found, and
89 questions produced multiple distinct solutions.

\section{Discussion}

A natural question is, ``did we faithfully replicate PRECISE?'' The
description of PRECISE was spread over two conference articles and a
couple of unpublished manuscripts. A forthcoming journal article was
referenced, but unfortunately it does not seem to have been
published. Several aspects of PRECISE were ambiguous, contradictory or
incomplete and forced us to make interpretations, which, if wrong,
could have an impact on recall. Still we made every effort to boost
evaluation results. For example, in section 2.4 we removed condition 4
from valid mappings and added the \underline{underlined} condition in
2. In section 2.2 we added the additional stop words and WH-words to
boost recall. Finally we omitted certain foreign keys from the lexicon
to limit needless ambiguity. We stand by the formalization presented
in section 2 as a reasonable interpretation of PRECISE, although we
are open to correction.

While the recall results did not replicate, at face value precision
results do appear to hold up; if one reads the questions under
reasonable interpretations, all the semantically tractable questions
map to what intuitively seems to be the correct SQL. Still one must
limit this claim. Consider that there is only one valid mapping for
the question ``what are the titles of films directed by George
Lucas?'', however a user may be disappointed if they expect the
database to also contain his student films. Similar misconceptions could
be present for attributes and values. This aside, our way to judge
correctness is based on common sense, assuming that the user fully
understands the context of the database. That said, \emph{the
  semantically tractable class does not seem to be fundamental}. We
have generalized the class and nothing seems to blocks the extension
of the class to questions requiring aggregation, superlatives,
negation, self-joins, etc. Also, the current semantically tractable
class excludes questions that seem simple (e.g. ``which films are
showing in los angeles?'' is not semantically tractable). Future work
is needed to more cleanly define and limit `semantically tractable'.

An issue that complicates PRECISE is the role of ambiguity. If the
user asks ``what are the titles of the Clint Eastwood films?'',
there are several possibilities: 1. The films he directed; 2. the
films he acted in; 3. the films he both acted and directed in; 4. the
films he either acted or directed in. Only 1 and 2 are expressible in
PRECISE. Still if there was a paraphrasing capability, the user could
select their intended interpretation. This leads to an immediate
strategy to improve practical 'recall'. Another immediate idea is to
extend PRECISE to handle ellipsis of WH-words.

A more serious issue is the hidden assumptions PRECISE makes about the
form of the schema. Natural language interfaces do better when the
schema maintains a clear relation with a conceptual model
(e.g. Entity-Relationship model). This is the case for example we
developed, but it is not completely the case for GEOQUERY which
contains tables such as \verb+HighLow+ which have no real entity
correspondence. Not surprisingly many of the rejected questions in our
evaluation involved this conceptually suspect table. What is needed is
a more specific delineation of exactly what schemas PRECISE is
applicable over. We shall look investigate this theoretically as well as
empirically, investigating for example how well PRECISE and it
generalizations cover QALD\cite{qald} and other corpora. 

\section{Conclusions}

Our replication of PRECISE made no errors in terms of returning a
single, incorrect query, giving it the highest possible precision
value. However, out of the 448 questions given, PRECISE was only able
to produce SQL queries for 162, giving it a recall value of
0.361. Moreover our implementation of PRECISE requires manual lexicon
configuration. Still, even given this `negative' result, we feel that
PRECISE is a very appealing approach, but one that needs more careful
scrutiny, testing and generalization. This is something we shall
continue to investigate.


\begin{thebibliography}{}

\bibitem[\protect\citename{Androutsopoulos, et. al.}2000]{androutsopoulos}
Ion Androutsopoulos and Graeme Ritchie.
\newblock {\em Database interfaces}.
\newblock In R.~Dale, H.~Moisl, and H.~Somers, editors, {\em Handbook of
  Natural Language Processing}, pages 209--240. Marcel Dekker Inc., 2000.

\bibitem[\protect\citename{Chu and Meng}1999]{meng99} Wesley Chu,
  and Frank Meng \newblock {\em Query Formulation from
    Natural Language using Semantic Modeling and Statistical Keyword
    Meaning Disambiguation} \newblock Technical Report 990003, UCLA CS
  Dept. 1999.

\bibitem[\protect\citename{De Marneffe, et. al.}2006]{stanford}
  Marie-Catherine De Marneffe, Bill MacCartney, Bill, and Manning,
  Christopher. {\em Generating typed dependency parses from phrase
  structure parses}. Proceedings of LREC 24, pages 449-454, 2006.

\bibitem[\protect\citename{Popescu,Etzioni and Kautz}2003]{precise03}
Ana-Maria Popescu, Oren Etzioni, and Henry Kautz. 
\newblock {\em Towards a theory of natural language interfaces to
databases} 
\newblock Proceedings of the 8th international conference on
Intelligent user interfaces 
\newblock pages 149-157, 2003.

\bibitem[\protect\citename{Popescu, et. al.}2004]{precise04} Ana-Maria
  Popescu, Alex Armanasu, Oren Etzioni, David Ko, and Alexander
  Yates. \newblock {\em Modern natural language interfaces to
    databases: Composing statistical parsing with semantic
    tractability.}  \newblock Proceedings of the 20th international
  conference on Computational Linguistics, \newblock 2004.

\bibitem[\protect\citename{Walter, et. al.}2012]{qald} Sebastian
  Walter, Christina Unger, Philipp Cimiano and Daniel B{\"a}r.  \newblock
  {\em Evaluation of a Layered Approach to Question Answering over
    Linked Data.} The Semantic Web - {ISWC} 2012 - 11th International
  Semantic Web Conference. pages 362--374. 2012.

\end{thebibliography}
\end{document}